\documentclass[letterpaper, 10 pt, conference]{ieeeconf}  

\IEEEoverridecommandlockouts                              

\usepackage{graphics} 
\usepackage{epsfig} 
\usepackage{amsmath} 
\usepackage{amssymb}  
\usepackage{makecell}
\usepackage{multirow}
\usepackage{soul}
\usepackage{color}
\usepackage{algorithm}
\usepackage{algpseudocode}
\usepackage{comment}
\usepackage{hyperref}

\hypersetup{pdfcreator={Me}}

\title{\LARGE \bf
Experience in Engineering Complex Systems: Active Preference Learning with Multiple Outcomes and Certainty Levels
}

\author{Le Anh Dao$^{1}$, Loris Roveda$^{2}$, Marco Maccarini$^{2}$, Matteo Lavit Nicora$^{1,3}$, Marta Mondellini$^{1}$,\\Matteo Meregalli Falerni$^{1}$, Palaniappan Veerappan$^{4}$, Lorenzo Mantovani $^{4}$, Dario Piga$^{2}$,\\Simone Formentin$^{4}$, Matteo Malosio$^{1}$
\thanks{*This work was partially supported by Regione Lombardia and Fondazione Cariplo in the framework of the project EMPATIA@Lecco—EMpowerment del PAzienTe In cAsa—Rif. 2016–1428, Decreto Regione Lombardia 6363 del 30/05/2017.}
\thanks{*This work was partially supported by the project ASSASSINN, funded from H2020 CleanSky 2 under grant agreement n. 886977.}
\thanks{*This work was partially supported by the project RoboPref, funded from Hasler Stiftung.}
\thanks{$^{1}$ Istituto di Sistemi e Tecnologie Industriali per il Manifatturiero Avanzato, Consiglio Nazionale delle Ricerche, via G. Previati 1/E, 23900 Lecco, Italia {\tt\small \{leanh.dao, matteo.lavit, marta.mondellini, matteo.malosio\}@stiima.cnr.it}}%
\thanks{$^{2}$ Istituto Dalle Molle di studi sull'Intelligenza Artificiale (IDSIA), Scuola Universitaria Professionale della Svizzera Italiana (SUPSI), Università della Svizzera Italiana (USI), Via la Santa 1, 6962 Lugano, Switzerland {\tt\small \{loris.roveda, marco.maccarini, dario.piga\}@idsia.ch}}%
\thanks{$^{3}$ Industrial Engineering Department, University of Bologna, 40131 Bologna, Italy {\tt\small matteo.lavitnicora2@unibo.it}}%
\thanks{$^{4}$ Politecnico di Milano, Italy {\tt\small \{palaniappan.veerappan, lorenzo1.mantovani\}@mail.polimi.it}, \tt\small \{simone.formentin\}@polimi.it}%
}

\begin{document}

\maketitle
\thispagestyle{empty}
\pagestyle{empty}

\begin{abstract}

Black-box optimization refers to the optimization problem whose objective function and/or constraint sets are either unknown, inaccessible, or non-existent. In many applications, especially with the involvement of humans, the only way to access the optimization problem is through performing physical experiments with the available outcomes being the preference of one candidate with respect to one or many others. Accordingly, the algorithm so-called Active Preference Learning has been developed to exploit this specific information in constructing a surrogate function based on standard radial basis functions, and then forming an easy-to-solve acquisition function which repetitively suggests new decision vectors to search for the optimal solution. Based on this idea, our approach aims to extend the algorithm in such a way that can exploit further information effectively, which can be obtained in reality such as: 5-point Likert type scale for the outcomes of the preference query (i.e., the preference can be described in not only  “this is better than that” but also  “this is much better than that” level), or multiple outcomes for a single preference query with possible additive information on how certain the outcomes are. The validation of the proposed algorithm is done through some standard benchmark functions, showing a promising improvement with respect to the state-of-the-art algorithm. 
\end{abstract}

\section{INTRODUCTION}

To find a solution of a real-world optimization problem, normally we formulate the corresponding problem into an explicit formulation and then find values of the decision variables to minimize the defined cost function or maximize a utility cost. But in many applications, the explicit mathematical expression of the objective function may either be expensive to obtain, or inaccessible, or not quantifiable due to the qualitative nature of the problem, or involve computer simulation process or laboratory experiments such that their inputs and outputs are available without analytical information of the inner working process \cite{Audet2017}. These problems are those, by definition, fall under the group of black-box optimization problems.

Generally, in these situations, the optimization problem can normally only be accessed through the evaluation of solution candidates through simulations or even physical experiments. One of the special cases is when humans are involved in the process of making assessment. For example, the involvement of humans in design and assessment process is important in health-related technologies \cite{mc2009user}: taking into account users’ needs may promote products' daily use \cite{thilo2017involvement} and facilitate long-term usage of health-related technologies \cite{facey2010patients}. Considering users’ needs and preferences follows the user-centered design philosophy \cite{norman1986user}. Users may be involved in designing autonomous mobile robots so that their behavior meets users' expectations \cite{Wilde2020}; or be involved in evaluating the fidelity motion of a wheelchair simulator for rehabilitation \cite{Dao2020}; or they are asked for feedback on new technologies' usability \cite{colombo2020mobile}.

Aligning with this context, a few authors have been studying in exploiting the outcomes of the preference query between two or more decision vectors for finding the optimal solution of the black-box optimization problem. Various works have been studied in last ten - twenty years, including \cite{Abdolshah2019} or \cite{Gonzalez2017} with Bayesian optimization algorithms, or in \cite{Ismael2007} with particle swarm optimization (PSO) algorithms. A recently developed algorithm, so called Active Preference Learning based on Radial Basis Function (APL- RBF), uses general radial basis functions (RBFs) to model the surrogate function of the latent one; in this work, the surrogate function is constructed in such a way to satisfy, if possible, the preferences already expressed by the decision maker at sampled points. At each iteration, the RBF weights are computed by solving a linear or quadratic programming problem, aiming to satisfy the available training set of pairwise preferences. Then, the proposed algorithm (denoted as AmPL - Active multi-Preference Learning) also constructs an acquisition function that is easy-to-evaluate and is minimized to generate a new sample and to query a new preference in the process of searching the global optimal solution. The approach has been validated and claimed to be, within the same number of comparisons, computationally lighter and approach the global optimum more closely than Bayesian active preference learning (PBO). With respect to PSO, the approach seems to conceive to the global optimal solution in a faster manner. Although the algorithm has been proven to effectively search for the global optimization in a reasonable number of tests, it is clear that the questionnaire used as well as the outcomes were rather simple and did not consider the uncertainty in the provided answer as it would happen in reality. Hence, it is natural to raise questions on whether and under which conditions a more complicated questionnaire could be inserted, and how this obtained information would be effectively exploited in the optimization problem. 

In the present study, some neuropsychological aspects must be considered for this premise to be satisfied. First of all, the human mind chooses more easily among few options, while it is in difficulty when facing many options \cite{bemporad2019active}: many choices can reduce one's satisfaction with the decision, the likelihood of making a decision and the decision's quality  \cite{besedevs2015reducing}. This difference has been observed also in users’ brain activation \cite{chau2014neural}. For these reasons, psychological studies suggest focusing on discrete choices as early as the 1920s \cite{bemporad2019active}.
On the other hand, marketing psychologists suggest that a wide range of choices allow subjects to find the option that best suits their taste \cite{baumol1956variety}, creates the perception of freedom of choice \cite{kahn1987experiments} and reinforces the overall satisfaction about the choice \cite{botti2004psychological}. Furthermore, the difficulty of choice is not only a direct consequence of the number of options but is mediated by various factors including time pressure \cite{haynes2009testing} and preference uncertainty \cite{chernev2015choice}.
In conclusion, limiting the number of possible choices and proposing a more varied option than “A is better than B” would seem to be the best option.
The present study considers also that the debate regarding the optimal number of choices in a Likert-type scale continues \cite{croasmun2011using}, \cite{joshi2015likert}, thus a 5-point Likert-type scale is used.
A final consideration is that most of the models that study the choice of preferences are based on the random utility theory, which states that subjects are always able to choose what they prefer  \cite{manski2001daniel};  however, this theory is not an accurate description of human behavior \cite{kelly2017think} and when people have to choose between similar options, the answers are very heterogeneous \cite{beck2013consistently}. To overcome this problem, economists and marketing psychologists added a question about the degree of certainty regarding the answer after each choice task during their tests; this operation seems useful in eliminating potential distortions \cite{johannesson1999calibrating}. In \cite{champ1997using}, a scale from 1 to 10 of certainty (very uncertain - very certain) was used, and similarly in \cite{ethier2000comparison} and \cite{vossler2003payment}. In these studies, the certainty of wanting to donate money or make a payment was assessed; but a high level of certainty did not always correspond to an effective payment action. In \cite{johannesson1998experimental}, degrees of certainty were divided into two (fairly sure/absolutely sure): this method was found to be more effective in understanding users' real intentions. As far as we know, studies that require the degree of certainty regarding a preference in interacting with a tool do not exist. Thus, previous research in other topics has been an inspiration for this study to insert a question on certainty with a limited number of choices ($N = 4$).

To this point, the innovative contribution is to design an approach based on Active Preference Learning to easily and effectively exploit more comprehensive, yet practical, information provided by human decision maker in the preference-based optimization problems. Not only that, by allowing multiple outcomes with uncertainty as the feedback of preference query, much more information could be harvested than in the case where only \textit{absolutely sure} answers are allowed. To the best of our knowledge, no other work has made this consideration in the same context before. In practice, the present study aims to find the best solution possible within the fewest trials: these aspects are of paramount importance, as physical experiments, especially the ones involving humans in the loop, are expensive in many situations by way of time consumption, the complexity of performed experiments or even participants' mental health and well-being. For these reasons, the effectiveness of the proposed algorithm (AmPL) is tested and validated through a benchmark function.

The rest of the paper is organized as follows: The description of the problem of interest is in Section II, which can be solved by the proposed algorithm explained in detail in Section III. Section IV is devoted to validating the proposed algorithm in benchmark functions. The paper is concluded with some remarks reported in Section V.

\section{Problem statement}
The problem of interest, referred to as the black-box optimization problem with constraints, can be defined as follows:
\begin{equation} \label{eq:black-box optimization}
 \begin{aligned}
x^* = argmin_{x}  f(x) 
\\ \text{ s.t.}\quad   l \leq x \leq u, g_i(x) \leq  0 
 \end{aligned}                   
\end{equation}
where $l$ and $u$ are vectors of the lower bound and the upper bound of the decision variables (or decision vector) $x \in \mathbb{R}^n$, respectively; $g_i: \mathbb{R}^n \rightarrow \mathbb{R}$ are further constraints on the decision vector $x$. Same assumption as in \cite{bemporad2019active} is made so the condition of $g_i(x)$ is easy to evaluate, so that, for example, in the case of inequality constraints $g_i(x)$ being linear, they can be described as: $g_i(x) = A_i x - b_i, A_i \in \mathbb{R}^{1 \times n}, b_i \in \mathbb{R}$. As discussed in the introduction, the objective function $f(x)$ exists but assumed to be  non-accessible. Formally, let $x_1$ and $x_2$ be two $n$-element decision vectors of the optimization problem in (\ref{eq:black-box optimization}), then the values of $f(x_1)$ and $f(x_2)$ are not accessible, but only their comparison in the form of discrete feedback outcome $p:\mathbb{R}^n \times \mathbb{R}^n \rightarrow \{-2, -1, 0, 1, 2 \}$ and their corresponding certainty level $c: \mathbb{R}^n \times \mathbb{R}^n \rightarrow \{1, 2, 3, 4 \} $, are. The overall preference function is then defined as:
\begin{equation} \label{eq:preference function}
\begin{aligned}
\pi: \mathbb{R}^n \times \mathbb{R}^n \rightarrow \{-2, -1, 0, 1, 2 \} \times \{1, 2, 3, 4 \} \\
\pi(x_1, x_2) = (p(x_1, x_2),c(x_1, x_2)) \\
\end{aligned}
\end{equation}
where
\begin{equation} \label{eq:p - preference function}
p(x_1, x_2) = 
    \begin{cases}
      -2 \quad \text{:} \; x_1 \text{``much better" than} \; x_2\\
      -1 \quad \text{:} \; x_1 \text{``slightly better" than} \; x_2\\
      \ \ 0 \quad \text{:} \; x_1 \text{``as good as"} \; x_2\\
      \ \ 1 \quad \text{:} \; x_2 \text{``slightly better" than} \; x_1\\
      \ \ 2 \quad \text{:} \; x_2 \text{``much better" than} \; x_1\\
    \end{cases} \\  
\end{equation}
and
\begin{equation} \label{eq:c - preference function}
c(x_1, x_2) = 
    \begin{cases}
      1 \quad \text{:} \; \text{not so sure}\\
      2 \quad \text{:} \; \text{quite sure}\\
      3 \quad \text{:} \; \text{sure } \\
      4 \quad \text{:} \; \text{absolutely sure } \\
    \end{cases} \\  
\end{equation}

The consideration of uncertainty in the answer of the human decision maker also suggests that multiple outcomes would be acquired from one single preference query. Given $q_h$ outcomes \{$(p_{h,1}, c_{h,1}), (p_{h,2}, c_{h,2}), ..., (p_{h,q_h}, c_{h,q_h})$\} with $p_{h,1} < p_{h,2} < ... < p_{h,q_h}$, are obtained from one certain preference query $\pi_{h}(x_1, x_2)$, where $h$ is index of the query, several assumptions are made as follows:

- If confused, the human should be confused only between similar outcomes, suggesting that if $p_{h,i}$ and $p_{h,j}, \forall i, j \in \mathbb{N}, 1 \leq i, j \leq q$ are in the list of outcomes, then any values between them and belong to the set of $\{-2, -1, 0, 1, 2\}$ should also be in the list of $q_h$ outcomes.

- If there exists a certainty level of one outcome being \textit{absolutely sure} then $q_h$ must be 1, implying that there is only one outcome from the corresponding preference query.

- $(p_{h,1}, p_{h,2},\dots, p_{h,q_h})$ should only take either non-negative values or non-positive values, suggesting that there is not a case where both ''$x_1$ is better than $x_2$" and ''$x_2$ is better than $x_1$" outcomes are provided by the human decision maker.

- While the human decision maker may not be \textit{absolutely sure} about individual outcomes, one of the outcomes needs to be the correct one.

Then, finding the optimal solution of the optimization problem in (\ref{eq:black-box optimization}) can be reinterpreted as finding $x^* \in \mathbb{R}^n$ such that:
\begin{equation} \label{eq:additiveConstraints}
p(x^*, x) \leq 0, \forall x: l \leq x \leq u \; \text{and} \; g_i(x) \leq 0  
\end{equation}

Although the objective function $f$ exists, there is no way to access it directly but we can only observe it through the outcomes $p(x_1, x_2)$ from the human decision maker. The outcomes imply the constraints on the difference between two decision vectors which can be mathematically formulated as:
\begin{equation} \label{eq:preference function mathematical form}
p(x_1, x_2) = 
    \begin{cases}
      -2 \quad \text{if} \; f(x_1) < f(x_2) - \sigma \\
      -1 \quad \text{if} \; f(x_2) - \sigma \leq f(x_1) < f(x_2) \\
      \ \ 0 \quad \text{if} \; f(x_1) = f(x_2) \\
      \ \ 1 \quad \text{if} \; f(x_2) + \sigma \geq f(x_1) > f(x_2) \\
      \ \ 2 \quad \text{if} \; f(x_1) > f(x_2) + \sigma \\
    \end{cases} \\  
\end{equation}
where $\sigma$ is an unknown positive number which is large enough to let the human decision maker perceive the significant difference between the values of the objective function with two solution candidates.

With the problem at hand, our goal of the proposed active preference learning algorithm is to suggest iteratively a sequence of samples $x_1 , x_2,. . . , x_N$ to test and compare such that $x_N$ approaches the optimal solution $x^*$ as $N$ grows.

\section{Active multi-Preference Optimization} 
This section is devoted to discussing in more detail the proposed active preference learning algorithm, denoted as AmPL (Active multi-Preference Learning), in searching for the optimal solution, exploiting as much of the information at hand as possible, including the ones with some level of uncertainty.
The proposed algorithm AmPL follows the same structure as the one discussed in \cite{bemporad2019active}: basically, the algorithm relies on two main steps of constructing the surrogate function and acquisition function. The surrogate function is a simpler-to-evaluate one to approximate the real objective function. On the other hand, to drive the search of new candidate to evaluate, the acquisition function is constructed whose optimal solution is the new decision vector to evaluate. These two main steps will be discussed in detail hereafter with respect to the new assumption considered. On the other hand, interested readers are invited to refer to \cite{bemporad2019active} and \cite{Bemporad2020} for other detail, such as generating the initial set of the decision vector using Latin hypercube sampling or tightening and scaling the decision vector and their constraints, and so on. In summary, the main steps of AmPL are reported in Algorithm 1 (Active multi-Preference Learning AmPL):

\begin{algorithm} [H]
\caption{Active multi-Preference Learning AmPL}\label{alg:cap}
\begin{algorithmic}[1]
\Require upper and lower bounds $(l, u)$, constraints set $g_i(x)$, hyper-parameters ($\sigma_1, \sigma_2, \alpha$), $N_{max}$ and $N_{init}$
\State \textbf{set}: $N \gets N_{init}$
\While{$N < N_{max}$}: \\
   Solve (\ref{eq:finding beta optimization}) for obtaining the surrogate function $\hat{f}$ \\
   Compute the acquisition function as in (\ref{eq:acquisition function}) and solve the global optimization problem to get $x_{N+1}$ \\
   Obtain the outcomes from possible preference queries between $x_{N+1}$ and any trials in the past (i.e.,$x_{1}$, $x_{2}$, \dots, $x_{N}$) that human decision maker still remembers\\
   Update $N$: $N \gets N+1$ 
\EndWhile
\end{algorithmic}
\end{algorithm}
$N_{max}$ is the maximum number of samples to evaluate; $N_{init}$ is number of initial samples.

\subsection{Preference-based surrogate function}
Since the objective function is assumed to be non accessible, to search for the optimal solution, a surrogate function of the objective function is built by actively learning from a finite set of sampled pairs of the decision vectors. 

Assuming that $N \geq 2$ samples ${x_1, x_2, ..., x_N}$ are generated with $x_i, x_j \in \mathbb{R}^n, x_i \neq x_j, \forall i, j = 1, 2, ..., N$. Then the preference vector $P = \{p_1, p_2, ..., p_M\}$ 
\begin{equation} \label{eq:preference_surrogate fucntion}
    p_h = p(x_{i(h)}, x_{j(h)}), p_h \in \{-2, -1, 0, 1, 2\}^{q_{h}}
\end{equation}
where $M$ is the number of pairwise comparison, $h = \{1,2,...,M\}, i(h) \ \text{and} \ j(h)$ are index of chosen pairs to evaluate, $i(h), j(h) \in \{1, 2, ..., N\}, i(h) \neq j(h)$, $q_h$ is the number of outcomes from the associated single preference query.

The goal is to find a surrogate function $\hat{f}: \mathbb{R}^n \rightarrow \mathbb{R}$ such that:
\begin{equation}\label{eq:objective_of_surrogate fucntion}
    p(x_{i(h)}, x_{j(h)}) = \hat{p}(x_{i(h)}, x_{j(h)})
\end{equation}
where $\hat{p}$ is preference function to the surrogate function $\hat{f}$ defining in the same way as in (\ref{eq:preference function mathematical form}). However, with the problem at hand, it happens sometimes that the $p(x_i(h), x_j(h))$ has several values of preference. In those cases, (\ref{eq:objective_of_surrogate fucntion}) becomes:
\begin{equation}\label{eq:objective_of_surrogate fucntion with uncertainty}
    \hat{p}(x_{i(h)}, x_{j(h)}) \in p(x_{i(h)}, x_{j(h)})
\end{equation}

As in \cite{bemporad2019active}, the surrogate function is built as the following linear combination of Radial Basis Functions (RBFs):
\begin{equation}\label{eq:surrogate_function_from_RBFs}
    \hat{f}(x) = \sum_{i=1}^{N}{\beta_i\phi(\gamma r(x, x_i))} 
\end{equation}
where the function $r: \mathbb{R}^{2n} \rightarrow \mathbb{R}$ is the Euclidean distance:
\begin{equation}\label{eq:Euclidean distance}
    r(x,y) = ||x-y||_2^2, x, y \in \mathbb{R}^n 
\end{equation}
$\phi: \mathbb{R} \rightarrow \mathbb{R}$ is a RBF whose value, by definition, depends only on the distance between input and some fixed points; $\gamma$ is a scalar hyper-parameter, which defines the shape of the RBF $\phi$, can be tuned through $K$-fold cross validation as in \cite{bemporad2019active}; $\beta_i$ are coefficients associating to RBF to construct the surrogate function, which can be solved through the following discussed convex optimization problem. While there are various types of the RBFs in literature, we have chosen the inverse quadratic type as \cite{bemporad2019active} so that the later comparison can be justified:
\begin{equation}
    \phi(\gamma r) = \frac{1}{1+(\gamma r)^2}
\end{equation}

With the aim to satisfy the objective referred in (\ref{eq:objective_of_surrogate fucntion}), (\ref{eq:objective_of_surrogate fucntion with uncertainty}), and following the imposed condition to the objective function $f$ in (\ref{eq:preference function mathematical form}), the following conditions are imposed to the surrogate function $\hat{f}$:
\begin{equation} \label{eq:surrogate function mathematical form}
    \begin{cases}
      p_{min, 0} = p_{h,1} < p_{h,1} <...< p_{h,q_h} = p_{max, 0}\\
      (*) \ m_{p_{min, 0}} \leq \hat{f}(x_{i(h)})-\hat{f}(x_{j(h)}) \leq M_{p_{max, 0}} \\
      (**) \ m_{p_{h,1},1} \leq \hat{f}(x_{i(h)})-\hat{f}(x_{j(h)}) \leq M_{p_{h,1},1} \\ 
      \vdots\\
      (**) \ m_{p_{h,q},q} \leq \hat{f}(x_{i(h)})-\hat{f}(x_{j(h)}) \leq M_{p_{h,q},q} \\ 
    \end{cases} \\ 
\end{equation}
where the lower and upper bounds in (\ref{eq:surrogate function mathematical form}) are set based on the values of their first subscripts $p_{h,1},..., p_{h,q}, p_{min, 0}, p_{max, 0} \in \{-2, -1, 0, 1, 2\}$ as follows:
\begin{equation} \label{eq:surrogate function mathematical form details}
    \begin{cases}
      m_{-2,t} &= -\infty; \ M_{-2,t} =  - \sigma_2 + \epsilon_{h,t} \\
      m_{-1,t} &= - \sigma_2 - \epsilon_{h,t}; \ M_{-1,t} =  - \sigma_1 + \epsilon_{h,t} \\
      m_{0,t} &=  - \sigma_1 - \epsilon_{h,t}; 
      \ M_{0,t} =  + \sigma_1 + \epsilon_{h,t} \\
      m_{1,t} &=  + \sigma_1 - \epsilon_{h,t}; \  
      M_{1,t} = + \sigma_2 + \epsilon_{h,t} \\
      m_{2,t} &=   + \sigma_2 - \epsilon_{h,t}, \ M_{2,t} = \infty \\
      &\forall t = 0, 1, 2, ..., q_h\\
    \end{cases} \\  
\end{equation}
in which $\sigma_1$ and $\sigma_2$ are given tolerances, and $\epsilon_{h,t}$ are slack variables. All of them take positive values. The slack variables allow one to relax the constraints imposed by the specified preferences $p_1, p_2, ..., p_M$. Constraint infeasibility might be due to an inappropriate selection of the RBF and/or to outliers in the acquired components $p_h$. The latter condition may easily happen when preferences $p_h$ are expressed by a human decision maker in an inconsistent way.
Some examples in the description in (\ref{eq:surrogate function mathematical form details}) are: (i) if the outcome of preference query in comparison between two decision vectors $x_{i(h)}$ and $x_{j(h)}$ is $p(x_{i(h)}, x_{j(h)}) = \{-1\}$ then $p_{max} = -1$ and $p_{min} = -1$; the corresponding conditions imposed to the surrogate function is $ - \sigma_2 - \epsilon_h \leq \hat{f}(x_{i(h)})-\hat{f}(x_{j(h)}) \leq - \sigma_1 + \epsilon_h$ or (ii) in another case, if multiple outcomes $p(x_{i(h)}, x_{j(h)}) = \{-2, -1, 0\}$ are collected by a single preference query then $p_{max} = 0$ and $p_{min} = -2$; the corresponding conditions imposed to the surrogate function is $-\infty \leq \hat{f}(x_{i(h)})-\hat{f}(x_{j(h)})  \leq + \sigma_1 + \epsilon_{h,0}$. Additional constraints (i.e., inequalities (**) in (\ref{eq:surrogate function mathematical form}) are: 
\begin{equation} \label{eq:additional constraints example}
    \begin{cases}
       p_{h,1} = -2: \\
       -\infty \leq \hat{f}(x_{i(h)})-\hat{f}(x_{j(h)}) \leq - \sigma_2 + \epsilon_{h,1} \\
       p_{h,2} = -1:\\
       - \sigma_2 - \epsilon_{h,2} \leq \hat{f}(x_{i(h)})-\hat{f}(x_{j(h)}) \leq \sigma_1 + \epsilon_{h,2} \\
       p_{h,3} = 0: \\
       - \sigma_1 - \epsilon_{h,3} \leq \hat{f}(x_{i(h)})-\hat{f}(x_{j(h)}) \leq \sigma_1 + \epsilon_{h,3} \\ 
    \end{cases} \\  
\end{equation}
Clearly, without the presence of slack variables $\epsilon_{h,1}, \epsilon_{h,2}, \epsilon_{h,3}$, these constraints do not overlap each other, and by minimizing any of these slack variables alone will force the term $\hat{f}(x_{i(h)})-\hat{f}(x_{j(h)})$ bounded inside the corresponding ranges. Since $\hat{f}(x_{i(h)})-\hat{f}(x_{j(h)})$ is a function of $\beta$ (see (\ref{eq:surrogate_function_from_RBFs})), as a result $\beta$ is indirectly affected by the constraints. Hence, the weighted sum of these slack variables will be inserted into the optimization problem  to find $\beta$; these weights' values depend on their associated certainty levels.

Accordingly, the parameters $\beta_i$, which define the form of $\hat{f}(x)$ as in (\ref{eq:surrogate_function_from_RBFs}), are obtained by solving the following convex optimization problem:
\begin{equation}\label{eq:finding beta optimization}
    \begin{aligned}
    \text{min}_{\beta_i, \epsilon_{h,0}, \epsilon_{h,t}} {\sum_{h=1}^M{b_h \epsilon_{h,0}}} +  \frac{\lambda}{2} {\sum_{k=1}^N{\beta_k^2}} + {\sum_{h=1}^M{\sum_{t=1}^{q_h}{w_{h,t} \epsilon_{h,t}}}}\\
    \text{s.t.} \ (\ref{eq:surrogate function mathematical form}) \text{ is satisfied}
    \end{aligned}
\end{equation}
where $b_h$ and $w_{h,t}$ are weights associated to the slack variable in (*) and (**) of (\ref{eq:surrogate function mathematical form}), respectively; $w_{h,t}$ values are selected proportionally to the certainty level $c_{h,t}$ to emphasize the fact that $\hat{f}(x_{i(h)})-\hat{f}(x_{j(h)})$ tends to stay inside a boundary where the certainty level is higher. In particular, $b_h = \sum_{t=1}^{q_h}{c_{h,t}}$, $w_{h,t}=\frac{c_{h,t}}{4}$ and $\lambda = 1$. Notice that  
in case $q_h = 1$ (i.e., only one outcome for the certain preference query) then the corresponding $w_{h,1} = 0$ as $m_{p_{min},0} = m_{p_{h,1},1}$ and $M_{p_{min},0} = M_{p_{h,1},1}$ then the roles of $\epsilon_{h,0}$ and $\epsilon_{h,1}$ are overlapped. Remind that the constraints on $\hat{f}(x_{i(h)})-\hat{f}(x_{j(h)})$ in (\ref{eq:surrogate function mathematical form}) can be rewritten as the function of $\beta_i$ following the description in (\ref{eq:surrogate_function_from_RBFs}).

\subsection{Preference-based acquisition function}
The procedure of finding the optimal decision vector based on the proposed active preference learning is to suggest iteratively a sequence of samples (i.e., decision vector) to test and compare such that $x_N$ approaches $x^*$ as N grows.
The acquisition is designed to do the work of suggesting a new sample by exploiting the surrogate function to suggest a new decision vector to evaluate. Since the surrogate function is built to approximate the objective function, a rather obvious option is to choose the acquisition function equal to the surrogate function. However, this selection of purely exploiting the surrogate function may lead to convergence to a point that is not the global optimization solution \cite{bemporad2019active}. Therefore, besides the \textit{exploitation} of the surrogate function, a function for \textit{exploration} must be taken into account to investigate other areas of the feasible space. Consider the exploration function $\mathbb{R}^n \rightarrow \mathbb{R}$ based on ideas from inverse distance weighting (IDW) as follows:
\begin{equation} \label{eq:exploration function}
z(x) = 
    \begin{cases}
       0     \ \ \ \    \text{if} \ x \in {x_1, ..., x_N}\\
       tan^{-1}(\frac{1}{\sum_{i=1}^N(1/{d^2(x, x_i)})})\\ 
    \end{cases} \\  
\end{equation}

With this function, we have $z(x_i) = 0 \ \forall x_i \in \{x_1, x_2, ..., x_N\}$ (i.e., the set of decision vectors already tested), $z(x) > 0, \forall x \in \mathbb{R}^n \setminus \{x_1, x_2, ..., x_N\}$, and $z(x)$ gets bigger as x being far away from all sampled points but assuring at the same time that $z(x)$ is refrained from getting excessively large. Then, the acquisition function $a: \mathbb{R}^n \rightarrow \mathbb{R}$ can be defined as:
\begin{equation} \label{eq:acquisition function}
a(x) = \frac{\hat{f}(x)}{\Delta \hat{F}} - \alpha_N z(x) \\
\end{equation}
where $\Delta \hat{F}$ is the range of the surrogate function on the sample list $\{x_1, x_2, ..., x_N\}$, which is used to normalize the $\hat{f}(x)$ to simplify the choice of exploration parameter $\alpha_N$; denoting $x^*_N$ as the \textit{current} best decision vector of sample list \{$x_1, x_2, ..., x_{N}$\} if $x^*_N$ is preferred to, or at least not worse than any decision vector  in the sample list \{$x_1, x_2, ..., x_{N}$\}. In AmPL, a version of varying $\alpha_N$ is studied, starting from a small value $\alpha_{min}$ after the new current best decision vector is found, to focus  more on exploiting the surrogate function but keeps growing to $\alpha_{max}$ to well explore the space of decision vectors until the next current best one is found:
\begin{equation} \label{eq:exploration weight}
\alpha_{N+1} = \\
    \begin{cases}
       0.2\Bar{\alpha}     \ \ \ \    \text{if} \ x_N = x^*_{N} \\
       \min(\alpha_{N} + 0.1\Bar{\alpha} , \Bar{\alpha}),  \ \text{otherwise}\\  
    \end{cases} \\  
\end{equation}
where $\bar{\alpha}$ are positive real values which are related to the exploration parameter. In APL-RBF, the exploration parameter $\alpha_N$ is fixed to $\bar{\alpha}, \forall N = \{1, 2, ..., N_{max}\}$.
At the end, the next sample $x_{N+1}$ is selected by solving:
\begin{equation} \label{eq:acquisition function solving}
 \begin{aligned}
x_{N+1} = argmin_{x}  a(x) 
\\ \text{ s.t.}\quad   l \leq x \leq b 
      \\ g_i(x) \leq  0 
 \end{aligned}  
\end{equation}

\section{Benchmark optimization problem}
\begin{figure} [ht]
    \centering
    \framebox{\parbox{3.2in}{
    \includegraphics[width=0.42\textwidth]{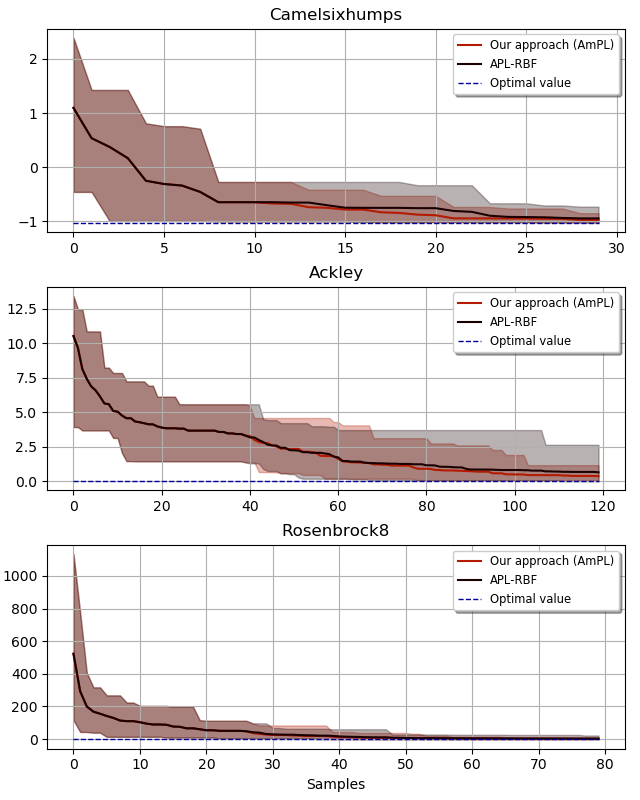}
}}
    \caption{Validation of the proposed algorithm - AmPL on the benchmark functions.}
    \label{fig:benchmark_validation}
\end{figure}

\begin{figure} [ht]
    \centering
    \framebox{\parbox{3.0in}{
    \includegraphics[width=0.40\textwidth]{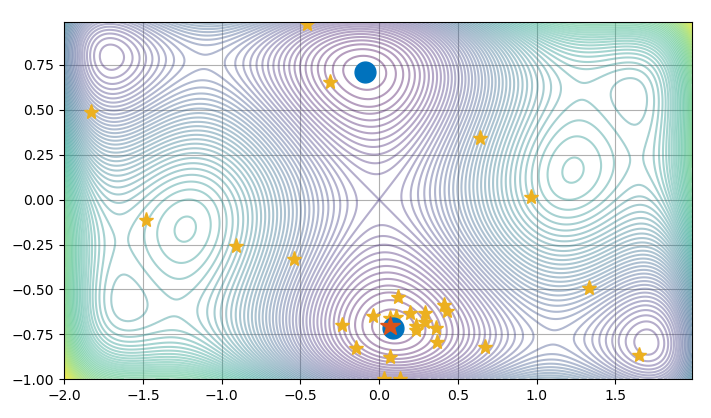}
}}
    \caption{Six-hump camel-back: Positions of the suggested samples using AmPL (yellow stars), of the best sample (red star), and of the optimal solution (blue circles).}
    \label{fig:Example_sample_position}
\end{figure}

In this section, AmPL is tested on several standard optimization problems which are included in \cite{Jamil2013}. The test functions comprise (i) the two-dimensional ``six-hump camel-back” (two global optimal solutions, six local minima), (ii) ``ackley" (typical multimodal test function), and the eight-dimensional ``Rosenbrock’s valley" (high demensional problem with the global optimum lays inside a long, narrow, parabolic shaped flat valley; denoted as Rosenbrock8) \cite{Molga2005}. Due to space limitation, the detailed description of these well-known objective functions will not be presented here but will follow the ones in \cite{Jamil2013} and their associated constraints in \cite{bemporad2019active}. 
For the sake of comparison, the benchmark problems are tested with our proposed algorithm - AmPL and APL-RBF with the same selection of the shared parameters and the initial sets of the decision vectors. The benchmark functions play the role of human decision maker in "answering" the preference query, and at the end to evaluate numerically AmPL in searching for the optimal solutions. As for AmPL, at each trial $N$, two preference values $p(x_N, x_{N-1})$ and $p(x_N, x^*_{N-1})$ are assessed, and each preference query provides maximum two outcomes with one of the two should be the correct one; also, their related certainty level from 1 to 4 will be randomly assigned to each outcome, while with the APL-RBF, only $p(x_N, x^*_{N-1})$ is assessed. Finally, the obtained results are reported in Fig. \ref{fig:benchmark_validation}. An example to show the positions of different samples suggested by AmPL and the optimal solution is depicted in Fig. \ref{fig:Example_sample_position}.

\begin{table}[h]
\caption{Validation with the benchmark functions: parameter settings and the obtained results}
\label{benchmark_parameter_setting}
\begin{center}
\begin{tabular}{|c|c|c|>{\bfseries}c|>{\bfseries}c||c|c|>{\bfseries}c|>{\bfseries}c|}
\hline
\multicolumn{9}{|c|}{Six-hump camel-back ($N_{init} = 10, N_{max} = 30$)} \\\hline
\multicolumn{5}{|c||}{AmPL} & \multicolumn{4}{c|}{APL-RBF} \\ \hline
\hline
$\bar{\alpha}$&$\sigma_1$&$\sigma_2$&$d_b$&$d_w$&$\bar{\alpha}$&$\sigma_1$&$d_b$&$d_w$ \\ \cline{1-9}
0.1&0.033&0.5&0.05&0.18&0.1&0.033&0.09&0.29\\ \hline

\multicolumn{9}{|c|}{Ackley ($N_{init} = 40, N_{max} = 120$)} \\
\hline
\multicolumn{5}{|c||}{AmPL} & \multicolumn{4}{c|}{APL-RBF} \\ \hline
$\bar{\alpha}$&$\sigma_1$&$\sigma_2$&$d_b$&$d_w$&$\bar{\alpha}$&$\sigma_1$&$d_b$&$d_w$ \\ \cline{1-9}
0.1&0.008&0.2&0.38&1.15&0.1&0.008&0.64&2.65\\\hline
\multicolumn{9}{|c|}{Rosenbrock8 ($N_{init} = 27, N_{max} = 80$)} \\
\hline
\multicolumn{5}{|c||}{AmPL} & \multicolumn{4}{c|}{APL-RBF} \\ \hline
$\bar{\alpha}$&$\sigma_1$&$\sigma_2$&$d_b$&$d_w$&$\bar{\alpha}$&$\sigma_1$&$d_b$&$d_w$ \\ \cline{1-9}
0.1&0.013&2&2.37&12.25&0.1&0.013&3.79&23.75\\ \hline

\end{tabular}
\end{center}
\end{table}
  
In Fig. \ref{fig:benchmark_validation}, besides the average performance of 20 different runs being shown, the band defined by the best- and worst-case instances obtained is also reported in the same figure. Numerical results of the distance between the optimal value and the obtained worst performance ($d_w$) and the mean of the best performances ($d_b$) obtained until the last tests of all runs are shown in Table \ref{benchmark_parameter_setting}. It is clear to see that AmPL significantly improves both the worst case and the mean performance in all the runs with the benchmark functions. 

\newpage
\section{Conclusions} 
This paper deals with a preference-based black-box optimization problem. A new, practical way of presenting the preference is proposed with the 5-point Likert-type scale together and their related certainty level. Since the certainty levels and a wide range of outcomes are involved, it is freer for the human decision maker to determine their feedback, and as a result, more information can be obtained. Starting from this consideration, the tests performed with the benchmark functions validated the efficiency of the proposed algorithm - AmPL with respect to the original APL - RBF algorithm.

Future work is devoted to testing the algorithm in experimental scenarios. Moreover, the authors intend to improve the approach so that it can easily handle further unused information, such as the subjective feeling on individual aspects of multi-objective optimization problem, or evaluate a decision vector itself without being compared to others, etc. Moreover, studies on choosing the suitable hyper-parameters or combining the knowledge of expert users and the algorithm in suggesting new samples are also interesting directions to investigate.



\end{document}